\documentclass{article}

\usepackage[final]{nips_2018}

\usepackage[utf8]{inputenc} 
\usepackage[T1]{fontenc}    
\usepackage{hyperref}       
\usepackage{url}            
\usepackage{floatflt}
\usepackage{lipsum}
\usepackage[multiple]{footmisc}
\usepackage{wrapfig}
\usepackage{booktabs}       
\usepackage{amsfonts}       
\usepackage{nicefrac}       
\usepackage{microtype}      
\usepackage{graphicx}
\usepackage{amsmath}
\usepackage{placeins}
\usepackage{natbib}
\usepackage{mathtools}
\usepackage{xcolor}
\usepackage{algorithm}
\usepackage{algorithmic}
\usepackage{color}
\usepackage{multirow}
\usepackage[font=small,labelfont=bf]{caption}

\definecolor{mydarkblue}{rgb}{0,0.08,0.45}

\def\mathunderline#1#2{\color{#1}\underbracket{{\color{black}#2}}\color{black}}

\def\mathunderline#1#2{\color{#1}\underbracket[1pt][1pt]{{\color{black}#2}}\color{black}}

\hypersetup{colorlinks=true,
    linkcolor=mydarkblue,
    citecolor=mydarkblue,
    filecolor=mydarkblue,
    urlcolor=mydarkblue}

\newcommand{\expect}{\mathbb{E}}
\newcommand{\transpose}{\top}
\newcommand{\kldiv}{\mathrm{D}_{\mathrm{KL}}}
\newcommand{\klbars}{\,\|\,}

\DeclareMathOperator*{\kvec}{vec}

\newcommand{\given}{\,|\,}
\newcommand{\normal}{\mathcal{N}}
\newcommand{\matrixNormal}{\mathcal{M} \mathcal{N}}
\newcommand{\ident}{\mathbf{I}}

\DeclareMathOperator*{\diag}{diag}

\newcommand{\data}{\mathcal{D}}
\newcommand{\elbo}{\mathcal{L}}

\newcommand{\lrate}{\alpha}
\newcommand{\lratePrec}{\beta}

\newcommand{\fisher}{\mathbf{F}}

\newcommand{\loss}{h}
\newcommand{\natgrad}{\tilde{\nabla}}
\newcommand{\cov}{{\boldsymbol{\Sigma}}}

\newcommand{\AMat}{\mathbf{A}}
\newcommand{\SMat}{\mathbf{S}}
\newcommand{\weightMatrix}{\mathbf{W}}
\newcommand{\activation}{\mathbf{a}}
\newcommand{\preactivation}{\mathbf{s}}

\newcommand{\mean}{{\boldsymbol \mu}}
\newcommand{\btheta}{{\boldsymbol \theta}}
\newcommand{\weights}{\mathbf{w}}
\newcommand{\qParams}{{\boldsymbol \phi}}
\newcommand{\klWeight}{\lambda}
\newcommand{\ndata}{N}
\newcommand{\target}{y}
\newcommand{\inputVec}{\mathbf{x}}
\newcommand{\priorVar}{\eta}
\newcommand{\lrateMod}{\tilde{\lrate}}
\newcommand{\lratePrecMod}{\tilde{\lratePrec}}

\newcommand{\grad}{\nabla}

\newcommand{\stdVec}{\boldsymbol{\sigma}}

\title{Eigenvalue Corrected Noisy Natural Gradient}

\author{
Juhan Bae${}^{1,2}$,
Guodong Zhang${}^{1,2}$,
Roger Grosse${}^{1, 2}$\\
University of Toronto${}^1$  Vector Institute${}^2$\\
\texttt{juhan.bae@mail.utoronto.ca} \\
\texttt{\{gdzhang, rgrosse\}@cs.toronto.edu}
}

\begin{document}

\maketitle

\begin{abstract}
Variational Bayesian neural networks combine the flexibility of deep learning with Bayesian uncertainty estimation. However, inference procedures for flexible variational posteriors are computationally expensive. A recently proposed method, noisy natural gradient, is a surprisingly simple method to fit expressive posteriors by adding weight noise to regular natural gradient updates. Noisy K-FAC is an instance of noisy natural gradient that fits a matrix-variate Gaussian posterior with minor changes to ordinary K-FAC. Nevertheless, a matrix-variate Gaussian posterior does not capture an accurate diagonal variance. In this work, we extend on noisy K-FAC to obtain a more flexible posterior distribution called eigenvalue corrected matrix-variate Gaussian. The proposed method computes the full diagonal re-scaling factor in Kronecker-factored eigenbasis. Empirically, our approach consistently outperforms existing algorithms (e.g., noisy K-FAC) on regression and classification tasks.
\end{abstract}

\section{Introduction}
Building flexible and scalable uncertainty models~\citep{mackay1992practical,neal2012bayesian,hinton1993keeping} has long been a goal in Bayesian deep learning. Variational Bayesian neural networks~\citep{graves2011practical, blundell2015weight} are especially appealing because they combine the flexibility of deep learning with Bayesian uncertainty estimation. However, such models tend to impose overly restricted assumptions (e.g., fully-factorized) in approximating posterior distributions. There have been attempts to fit more expressive distributions~\citep{louizos2016structured, sun2017learning}, but they are difficult to train due to strong and complicated posterior dependencies.

\begin{wrapfigure}[13]{R}{0.35\textwidth}
    \vspace{-1.4em}
    \centering
    \includegraphics[width=1.8in]{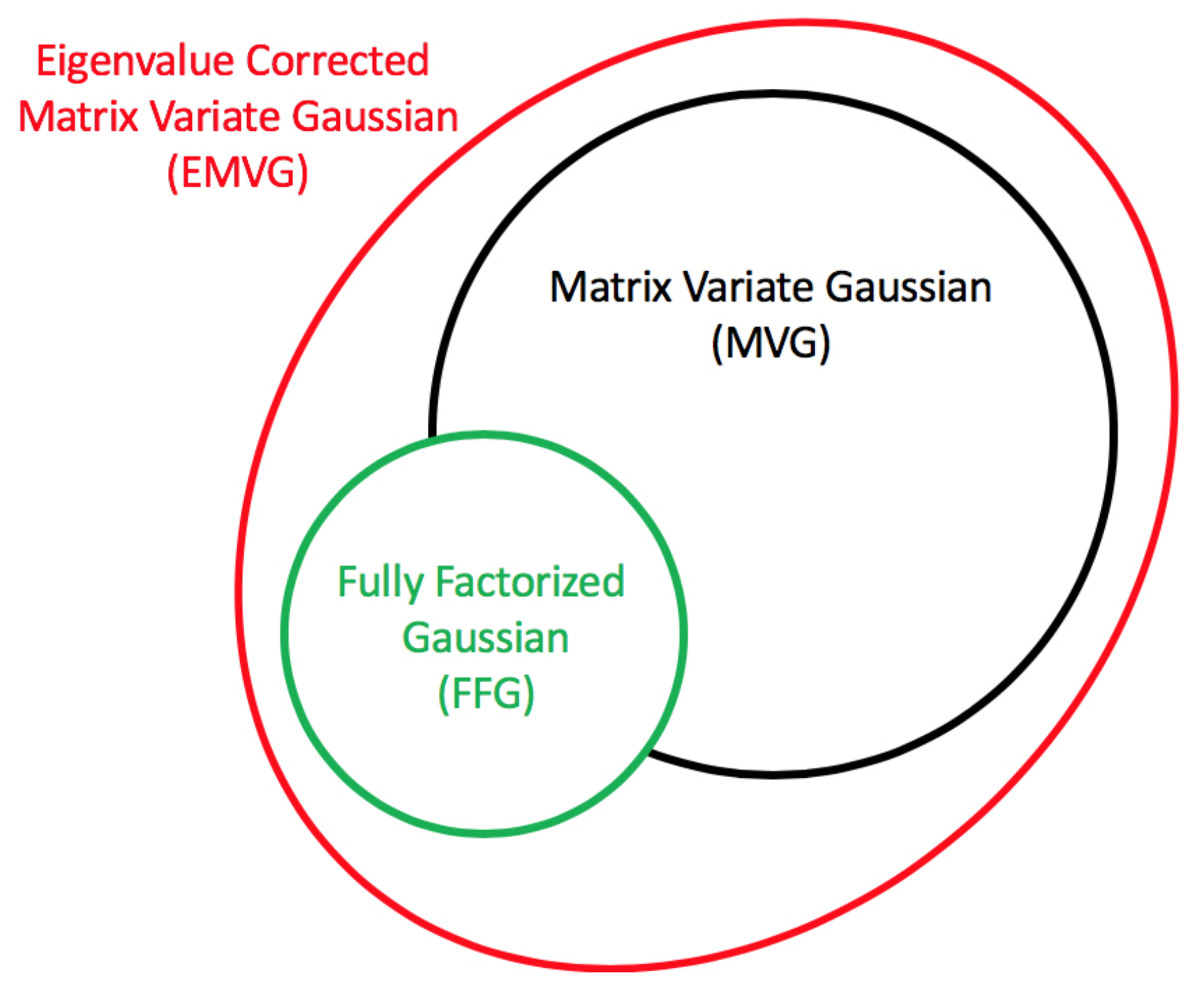}
    \small
    \caption{A cartoon illustration to describe the relationships of FFG, MVG, and EMVG.}
    \label{fig:cartoon-fig1}
\end{wrapfigure}

Noisy natural gradient is a simple and efficient method to fit multivariate Gaussian posteriors~\citep{zhang2017noisy}. It adds adaptive weight noise to regular natural gradient updates. Noisy K-FAC is a practical algorithm in the family of noisy natural gradient~\citep{zhang2017noisy}, which fits a matrix-variate Gaussian posterior (flexible posterior) with only minimal changes to ordinary K-FAC update~\citep{martens2015optimizing} (cheap inference). The update for noisy K-FAC closely resembles standard K-FAC update with correlated weight noise. 

Nevertheless, we note that a matrix-variate Gaussian cannot capture an accurate diagonal variance. In this work, we build upon the large body of noisy K-FAC and Eigenvalue corrected Kronecker-factored Approximate Curvature (EK-FAC)~\citep{george2018fast} to improve the flexibility of the posterior distribution. We compute the diagonal variance, not in parameter coordinates, but in K-FAC eigenbasis. This leads to a more expressive posterior distribution. The relationship is described in Figure~\ref{fig:cartoon-fig1}. Using this insight, we introduce a modified training method for variational Bayesian neural networks called \emph{noisy EK-FAC}. 

\section{Natural Gradient}
\label{background}
\begin{figure}[t]
    \centering
    \includegraphics[width=0.8\textwidth]{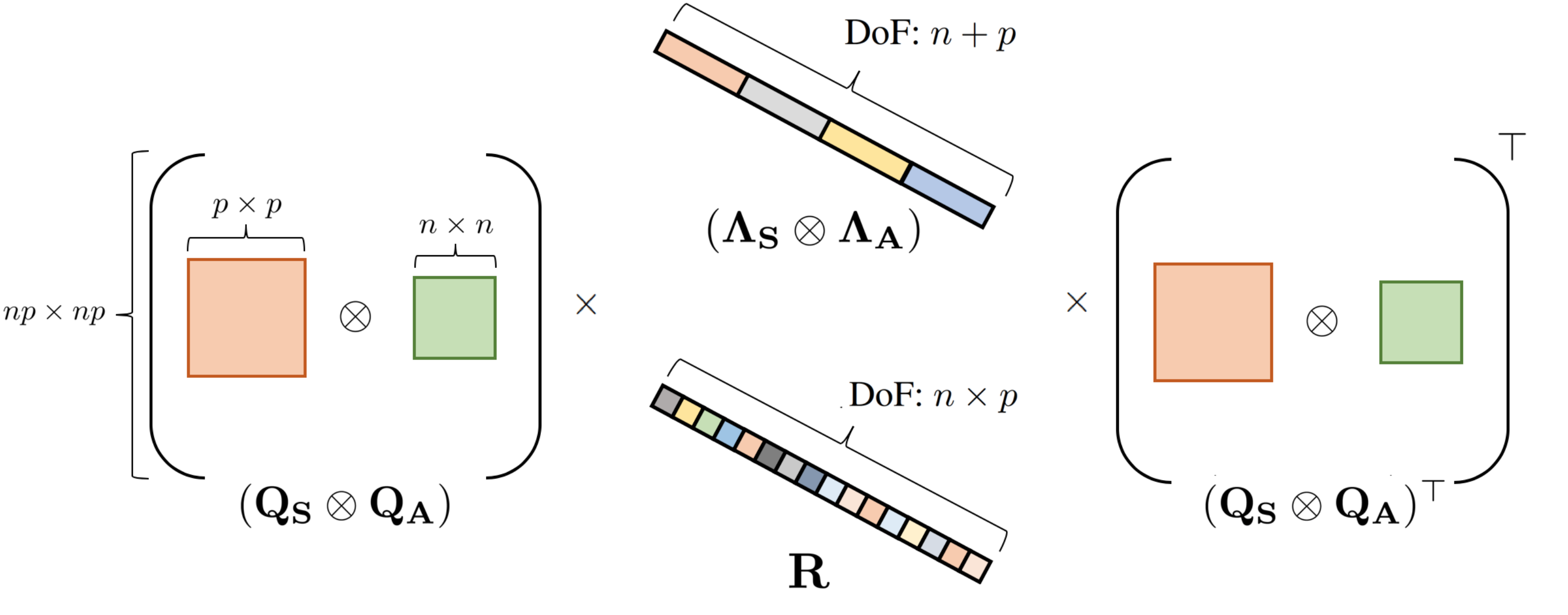}
    \caption{The diagonal re-scaling factor in K-FAC has Kronecker structure with \(n + p\) degrees of freedom. The diagonal re-scaling matrix in EK-FAC is the second moment of the gradient vector with \(n \times p\) degrees of freedom.}
    \label{fig:diff_ekfac}
\end{figure}

Natural gradient descent is a second-order optimization technique first proposed by~\citet{amari1997neural}. It is classically motivated as a way of implementing steepest descent in the space of distributions instead of the space of parameters. The distance function for distribution space is the KL divergence on the model's predictive distribution: \(\kldiv(p_\mathbf{\btheta}\klbars p_{\mathbf{\btheta} + \Delta\mathbf{\btheta}}) \approx \frac{1}{2}\mathbf{\btheta}^\transpose\fisher\mathbf{\btheta}\), where \(\fisher\) is the Fisher matrix.
\begin{equation}
    \centering
    \fisher = \expect \left[ \nabla_{\weights}\log p(\target | \inputVec, \weights) \nabla_{\weights}\log p(\target | \inputVec, \weights)^\top \right]
\end{equation}
This results in the preconditioned gradients \(\natgrad_\weights \loss = \fisher^{-1} \nabla_\weights \loss\). Natural gradient descent is invariant to smooth and invertible reparameterizations of the model~\citep{martens2014new}. 

\subsection{Kronecker-Factored Approximate Curvature}
Modern neural networks contain millions of parameters which makes storing and computing the inverse of the Fisher matrix impractical. Kronecker-Factored Approximate Curvature (K-FAC)~\citep{martens2015optimizing} uses Kronecker products to efficiently approximate the inverse Fisher matrix\footnote{Extending on this work, K-FAC was shown to be amenable to distributed computation~\citep{ba2016distributed} and could generalize as well as SGD~\citep{zhang2018three}.}. 

For a layer of a neural network whose input activations are \(\mathbf{a} \in \mathbb{R}^{n}\), weight matrix \(\weightMatrix \in \mathbb{R}^{n \times p}\), and pre-activation output \(\preactivation \in \mathbb{R}^{p}\), we can write \(\preactivation = \weightMatrix^\top\activation\). The gradient with respect to the weight matrix is \(\nabla_{\weightMatrix}\loss = \activation(\nabla_{\preactivation} \loss )^\top\). Assuming  \(\mathbf{a}\) and \(\mathbf{s}\) are independent under the model's predictive distribution, K-FAC decouples the Fisher matrix \(\mathbf{F}\):
\begin{equation}
\begin{aligned}
	\fisher &= \expect \left[\mathrm{vec}\{\grad_{\weightMatrix} \loss \}\mathrm{vec}\{\grad_{\weightMatrix}\loss \}^\top \right] = \expect \left[\{\grad_{\preactivation} \loss \}\{\grad_{\preactivation} \loss\}^\top \otimes \activation\activation^\top \right] \\
    &\approx \expect \left[\{\grad_{\preactivation} \loss\}\{\grad_{\preactivation} \loss \}^\top \right] \otimes \expect \left[\activation\activation^\top \right] = \mathbf{S} \otimes \mathbf{A}
\end{aligned}
\end{equation}
where \(\mathbf{A} = \expect \left[\activation\activation^\top \right]\) and \(\mathbf{S} = \expect \left[\{\grad_\preactivation \loss \}\{\grad_\preactivation \loss \}^\top \right]\).~\citet{bernacchiaexact} showed that the Kronecker approximation is exact for deep linear networks, justifying the validity of the above assumption. Further assuming the between-layer independence, the Fisher matrix is approximated as block diagonal consisting of layer-wise Fisher matrices. Decoupling \(\fisher\) into \(\mathbf{A}\) and \(\mathbf{S}\) avoids the memory issue of storing the full matrix \(\fisher\) while also having the ability to perform efficient inverse Fisher vector products:
\begin{equation}
	\fisher^{-1}\mathrm{vec}\{\grad_{\weightMatrix}\loss\} = \mathbf{S}^{-1} \otimes \mathbf{A}^{-1} \mathrm{vec}\{\grad_{\weightMatrix}\loss\} = \mathrm{vec}[\mathbf{A}^{-1} \grad_{\weightMatrix}\loss \mathbf{S}^{-1}]
	\label{eq:ng-kfac}
\end{equation}
As shown in equation~\eqref{eq:ng-kfac}, natural gradient descent with K-FAC only consists of a series of matrix multiplications comparable to the size of \(\weightMatrix\). This enables an efficient computation of a natural gradient descent.

\subsection{An Alternative Interpretation of Natural Gradient}
\label{new-interp-sec}
\citet{george2018fast} suggest an alternative way of interpreting the natural gradient update. It can be broken down into three stages:
\definecolor{aqua}{rgb}{0.0, 1.0, 1.0}
\definecolor{amber}{rgb}{1.0, 0.75, 0.0}
\begin{equation}
    \mathbf{F}^{-1}\mathrm{vec}\{\grad_{\weightMatrix}\loss\} = \mathunderline{amber}{\mathbf{Q}\mathunderline{aqua}{\mathbf{R}^{-1}\mathunderline{red}{\mathbf{Q}^\transpose\mathrm{vec}\{\grad_{\weightMatrix}\loss\}}}}
    \label{eq:new-inter}
\end{equation}
The first stage (\textcolor{red}{\textbf{--}}) projects the gradient vector to the full Fisher eigenbasis \(\mathbf{Q}\). The next step (\textcolor{aqua}{\textbf{--}}) re-scales the coordinates in full Fisher eigenbasis \(\mathbf{Q}\) with the diagonal re-scaling factor \(\mathbf{R}\). The last stage (\textcolor{amber}{\textbf{--}}) projects back to the parameter coordinates. 

For a diagonal approximation of the Fisher matrix, the basis is chosen to be identity matrix \(\ident\) and the re-scaling factor \(\mathbf{R}_{ii} = \mathbb{E}[(\nabla_{\weights})^2_i]\) is the second moment of the gradient vector. While estimating the diagonal factor is simple and efficient, obtaining an accurate eigenbasis is difficult. The crude basis in the diagonal Fisher introduces a significant approximation error.

K-FAC decouples the Fisher matrix into \(\mathbf{S}\) and \(\mathbf{A}\). Since \(\mathbf{S}\) and \(\mathbf{A}\) are symmetric positive semi-definite matrices, by eigendecomposition, they can be represented as \(\mathbf{S} = \mathbf{Q}_{\SMat} \mathbf{\Lambda}_{\SMat} \mathbf{Q}_{\SMat}^\top\) and \(\mathbf{A} = \mathbf{Q}_{\AMat} \mathbf{\Lambda}_{\AMat} \mathbf{Q}_{\AMat}^\top\), where \(\mathbf{Q}\) is an orthogonal matrix whose columns are eigenvectors and \(\mathbf{\Lambda}\) is a diagonal matrix with eigenvalues. We use properties of the Kronecker product to further decompose the factorization:
\begin{equation}
    \mathbf{S} \otimes \mathbf{A}= \mathbf{Q}_{\SMat} \mathbf{\Lambda}_{\SMat} \mathbf{Q}_{\SMat}^\transpose \otimes \mathbf{Q}_{\AMat} \mathbf{\Lambda}_{\AMat} \mathbf{Q}_{\AMat}^\transpose = (\mathbf{Q}_{\SMat} \otimes \mathbf{Q}_{\AMat})(\mathbf{\Lambda}_{\SMat} \otimes \mathbf{\Lambda}_{\AMat})(\mathbf{Q}_{\SMat} \otimes \mathbf{Q}_{\AMat})^\transpose
    \label{eq:eigen-fisher}
\end{equation}
Based on the new interpretation, we have K-FAC eigenbasis \(\mathbf{Q}_{\SMat} \otimes \mathbf{Q}_{\AMat}\) and diagonal re-scaling factor \(\mathbf{\Lambda}_{\SMat} \otimes \mathbf{\Lambda}_{\AMat}\). K-FAC eigenbasis is provably a more accurate approximation of the full Fisher eigenbasis. However, it does not use the estimated variance along the basis. The re-scaling factor in K-FAC is constrained to the Kronecker structure.

\subsection{Eigenvalue Corrected Kronecker-Factored Approximate Curvature}

Eigenvalue corrected K-FAC (EK-FAC)~\citep{george2018fast} extends on K-FAC to compute a more accurate diagonal re-scaling factor in K-FAC eigenbasis. The re-scaling factor for K-FAC is expressed in \(n + p\) degrees of freedom, where \(n\) and \(p\) are input and output size of a layer. K-FAC factorization in equation~\eqref{eq:eigen-fisher} does not capture an accurate diagonal re-scaling factor in K-FAC eigenbasis because of the Kronecker structure. Instead, EK-FAC computes the second moment of the gradient vector in K-FAC eigenbasis. We define the re-scaling matrix \(\mathbf{R} \in \mathbb{R}^{np \times np}\) as follows: 
\begin{equation}
    \mathbf{R}_{ii} = \mathbb{E}[((\mathbf{Q}_{\SMat} \otimes \mathbf{Q}_{\AMat})^\transpose \nabla_{\weights})^2_i]
    \label{eq:re-scaling}
\end{equation}
\(\mathbf{R}\) is a diagonal matrix whose entries are the second moment. The Fisher matrix can be approximated with K-FAC eigenbasis and the re-scaling matrix:
\begin{equation}
    \fisher \approx (\mathbf{Q}_{\SMat} \otimes \mathbf{Q}_{\AMat})\mathbf{R}(\mathbf{Q}_{\SMat} \otimes \mathbf{Q}_{\AMat})^\transpose
    \label{eq:fisher-approx}
\end{equation}
EK-FAC re-scaling matrix minimizes the approximation error of the above equation in Frobenius norm. In comparison to K-FAC approximation, Eigenvalue corrected K-FAC (EK-FAC) approximation is more flexible in representing the diagonal re-scaling factor with \(n \times p\) degrees of freedom. Figure~\ref{fig:diff_ekfac} illustrates the difference between K-FAC and EK-FAC. 

\section{Variational Bayesian Neural Networks}
Given a dataset \(\mathcal{D} = \{(\mathbf{x}_i, y_i)^n_{i=1}\}\), a Bayesian Neural Network (BNN) is composed of a log-likelihood \(p(\mathcal{D} | \mathbf{w})\) and a prior \(p(\mathbf{w})\) on the weights. Performing inference on BNN requires integrating over the intractable posterior distribution $p(\weights\given\data)$.
Variational Bayesian methods~\citep{hinton1993keeping,graves2011practical,blundell2015weight} attempt to fit an approximate posterior \(q(\weights)\) to maximize the evidence lower bound (ELBO):
\begin{equation}
    \elbo[q] = \expect_{q}[\log p(\data \given \weights)] - \klWeight \kldiv(q(\weights)\klbars p(\weights))
\end{equation}
where \(\lambda\) is a regularization parameter and \(\phi\) are parameters of the variational posterior. The exact Bayesian inference uses \(\lambda = 1\), but it can be tuned in practical settings.

Bayes By Backprop (BBB)~\citep{blundell2015weight} is the most common variational BNN training method. It uses a fully-factorized Gaussian approximation to the posterior i.e.~\(q(\weights) = \normal(\weights;\mean, \diag(\stdVec^2))\).
The variational parameters \(\qParams = (\mean, \stdVec^2)\) are updated according to stochastic gradients of \(\elbo\) obtained by the reparameterization trick~\citep{kingma2013auto}. 

There has been attempts to fit a matrix-variate Gaussian posterior for BNNs~\citep{louizos2016structured, sun2017learning}. Compared to overly restricted variational families, a matrix-variate Gaussian effectively captures correlations between weights. However, computing the gradients and enforcing the positive semi-definite constraint for \(\mathbf{\Sigma}_1\) and \(\mathbf{\Sigma}_2\) make the inference challenging. Existing methods typically impose additional structures such as diagonal covariance~\citep{louizos2016structured} or products of Householder transformation~\citep{sun2017learning} to ensure efficient updates.

\subsection{Noisy Natural Gradient}
Noisy natural gradient (NNG) is an efficient method to fit multivariate Gaussian posteriors~\citep{zhang2017noisy} by adding adaptive weight noise to ordinary natural gradient updates\footnote{\citet{khan2018fast} also found the relationship
between natural gradient and variational inference and
derived VAdam by adding weight noise to Adam, which is similar to noisy Adam in~\citet{zhang2017noisy}.}. Assuming \(q(\weights)\) is a multivariate Gaussian posterior parameterized by \(\boldsymbol{\phi} = \left(\boldsymbol{\mu}, \cov \right)\) and \(p(\weights)\) is a spherical Gaussian, the update rules are:
\begin{equation}
  \begin{aligned}
    & \fisher \leftarrow \left(1 - \lratePrecMod\right) \fisher+\lratePrecMod \left(\nabla_{\mathbf{w}} \log p(y | \mathbf{x}, \mathbf{w})
\nabla_{\mathbf{w}} \log p(y | \mathbf{x}, \mathbf{w})^{\transpose}\right) \\
    &\boldsymbol{\mu} \leftarrow \boldsymbol{\mu} + \lrateMod \left(\mathbf{\fisher} + \frac{\klWeight}{\ndata \priorVar}\mathbf{I}\right)^{-1}\left(\nabla_{\mathbf{w}}\log p(y | \mathbf{x}, \mathbf{w}) - \frac{\lambda}{N \eta}\mathbf{w}\right) 
\label{eq:noisy-update}
\end{aligned}
\end{equation}
where \(\lambda\) is the KL weight and \(\eta\) is the prior variance. In each iteration, NNG samples weights from the variational posterior \(q_\phi(\weights)\), which is a multivariate Gaussian with the covariance matrix:
\begin{equation}
    \cov = \left(\frac{\ndata}{\klWeight} \fisher + \priorVar^{-1}\mathbf{I}\right)^{-1}
    \label{eq:posterior-variance}
\end{equation}
However, due to computational intractability, it is necessary to impose a structured restriction to the covariance matrix. This is equivalent to imposing the same structure to the Fisher matrix. 



\subsection{Fitting Matrix-Variate Gaussian Posteriors with Noisy K-FAC}
Noisy K-FAC is a tractable instance of NNG with Kronecker-factored approximation to the Fisher. Because imposing a structured approximation to the covariance is equivalent to imposing the same structure to the Fisher matrix, noisy K-FAC enforces a Kronecker product structure to the covariance matrix. It efficiently fits the matrix-variate Gaussian posterior. The posterior covariance is given by
\begin{equation}
    \begin{aligned}
    \cov &= (\mathbf{Q}_\mathbf{S} \otimes \mathbf{Q}_\mathbf{A})\left(\frac{N}{\lambda}(\mathbf{\Lambda}_{\SMat} \otimes \mathbf{\Lambda}_{\AMat}) + \eta^{-1}\ident\right)^{-1}(\mathbf{Q}_\mathbf{S} \otimes \mathbf{Q}_\mathbf{A})^\transpose \\
    &\approx \frac{\lambda}{N}\left(\mathbf{S} + \frac{1}{\pi} \sqrt{\frac{\lambda}{N\eta}}\mathbf{I}\right)^{-1} \otimes \left(\mathbf{A} + \pi \sqrt{\frac{\lambda}{N\eta}}\mathbf{I}\right)^{-1}
    \end{aligned}
\label{noisy-kfac}
\end{equation}
where \(\pi \) is a scalar constant introduced by~\citet{martens2015optimizing} in the context of damping to keep a compact representation of the Kronecker product. The pseudo-code for noisy K-FAC is given in Appendix~\ref{app:noisy-kfac}. In comparison to existing methods that fit MVG posteriors~\citep{sun2017learning, louizos2016structured}, noisy K-FAC does not assume additional approximations.

\section{Methods}
\label{Methods}
\begin{algorithm}[t]
\caption{Noisy EK-FAC.
A subscript \(l\) denotes the index of a layer, \(\weights_l = \kvec(\mathbf{W}_l)\), and \(\mean_l = \kvec(\mathbf{M}_l)\). We assume zero momentum for simplicity. \(\oslash\) denotes element-wise division and \(\mathbf{r}_l^\gamma = \textrm{diag}(\mathbf{R}_l^{\gamma})\). unvec\((\cdot)\) is an inverse of vec\((\cdot)\) operation i.e. \(\textrm{vec}(\mathbf{A}) = \mathbf{a}\) and \(\textrm{unvec}(\mathbf{a}) = \mathbf{A}\). Differences from standard EK-FAC are shown in {\color{blue} blue}.}
\label{alg:noisy-kfac}
\begin{algorithmic}
\REQUIRE \(\alpha\): stepsize
\REQUIRE \(\beta\): exponential moving average parameter for covariance factors
\REQUIRE \(\omega\): exponential moving average parameter for re-scaling factor
\REQUIRE \(\klWeight, \eta, \gamma_{\text{ex}}:\) KL weighting, prior variance, extrinsic damping term
\REQUIRE \(T_{\rm stats}, T_{\rm eig}, T_{\rm scale}\): stats, eigendecomposition, and re-scaling update intervals.
\STATE \(k \leftarrow 0\) and initialize \(\{\mean_l\}_{l=1}^{L}, \{\mathbf{S}_l\}_{l=1}^{L}, \{\mathbf{A}_l\}_{l=0}^{L-1}, \{\mathbf{R}_l\}_{l=1}^{L}\)
\STATE Calculate the intrinsic damping term \({\color{blue} \gamma_{\text{in}}=\frac{\klWeight}{\ndata \eta}}\), total damping term \({\color{blue} {\gamma=\gamma_{\text{in}}+\gamma_{\text{ex}}}}\)
\WHILE{stopping criterion not met}
	\STATE \(k \leftarrow k+1\)
        \STATE {\color{blue} {\(\mathbf{W}_l \sim \mathcal{E}\mathcal{M}\mathcal{N}(\mathbf{M}_l,\  \mathbf{A}_l,\ \mathbf{S}_l,\ \frac{\klWeight}{N}[\mathbf{R}_l^{\gamma_{\text{in}}}]^{-1})\)}}
        \IF {$k \equiv 0$ (\textrm{mod} $T_{\rm stats}$)}
	    \STATE Update the covariance factors
	    $\{\mathbf{S}_l\}_{l=1}^{L}, \{\mathbf{A}_l\}_{l=0}^{L-1}$ using eq.~\eqref{eq:moving-average}
        \ENDIF
        \IF {$k \equiv 0$ (\textrm{mod} $T_{\rm scale}$)}
	    \STATE Update the re-scaling factor  \(\{\mathbf{R}_l\}_{l=1}^{L}\) using eq.~\eqref{eq:moving-average}
        \ENDIF
        \IF {$k \equiv 0$ (\textrm{mod} $T_{\rm eig}$)}
	    \STATE Compute the eigenbasis \(\mathbf{Q}_{\SMat_l}\) and \(\mathbf{Q}_{\AMat_l}\) using eq.~\eqref{eq:eigen-fisher}. 
        \ENDIF
        \STATE \(\mathbf{V}_l = \nabla_{\mathbf{W}_l}\log p(\target | \inputVec, \weights) {\color{blue} \,-\, \gamma_{\text{in}} \cdot \mathbf{W}_l}\)
        \STATE \(\mathbf{M}_l \gets \mathbf{M}_l + \alpha
        \mathbf{Q}_{\AMat_l}[
        (\mathbf{Q}_{\AMat_l}^\transpose\mathbf{V}_l \mathbf{Q}_{\SMat_l}) \oslash \textrm{unvec}(\mathbf{r}_l^\gamma)]\mathbf{Q}_{\SMat_l}^\transpose\)
        \COMMENT{Derivation is shown in Appendix~\ref{app:derivation}}
\ENDWHILE
\end{algorithmic}
\end{algorithm}
While matrix-variate Gaussian posterior efficiently captures correlations between different weights, the diagonal variance in K-FAC eigenbasis is not optimal. K-FAC diagonal re-scaling factor \(\mathbf{\Lambda}_{\SMat} \otimes \mathbf{\Lambda}_{\AMat}\) does not match the second moment along the associated eigenvector  \(\mathbb{E}[((\mathbf{Q}_{\SMat} \otimes \mathbf{Q}_{\AMat})^\transpose \nabla_{\weights})^2_i]\). 

We develop a new tractable instance of noisy natural gradient. It keeps track of the diagonal variance in K-FAC eigenbasis, resulting in a more flexible posterior distribution. In the context of NNG, imposing a structural restriction to the Fisher matrix \(\fisher\) is equivalent to imposing the same restriction to the variational posterior. For example, noisy K-FAC imposes a Kronecker product structure to the covariance matrix as shown in equation~\eqref{noisy-kfac}.

Given these insights, building a flexible variational posterior boils down to finding an improved approximation of the Fisher matrix. We adopt EK-FAC method, which is provably a better approximation of the Fisher matrix than K-FAC. We term the new BNN training method \emph{noisy EK-FAC}.

EK-FAC uses eigenvalue corrected Kronecker-factored approximation to the Fisher matrix as described in equation~\eqref{eq:fisher-approx}. For each layer, it estimates \(\AMat, \SMat\), and \(\mathbf{R}\) online using exponential moving averages. Conveniently, this resembles the exponential moving average updates for the noisy natural gradient in equation~\eqref{eq:noisy-update}.
\begin{equation}
\begin{aligned}
    {}
    \AMat & \leftarrow (1 - \lratePrecMod) \AMat + \lratePrecMod \mathbf{a} \mathbf{a}^\transpose \\
    \SMat & \leftarrow (1 - \lratePrecMod) \SMat + \lratePrecMod \nabla_{\preactivation}\log p(\target | \inputVec, \weights)\nabla_{\preactivation}\log p(\target | \inputVec, \weights)^\transpose \\
    \mathbf{R}_{ii} & \leftarrow (1 - \tilde{\omega})\mathbf{R}_{ii} + \tilde{\omega} \left[(\mathbf{Q}_{\SMat} \otimes \mathbf{Q}_{\AMat})^\transpose \nabla_{\weights}\log p(\target | \inputVec, \weights)\right]_i^2
\end{aligned}
\label{eq:moving-average}
\end{equation}
where \(\lratePrecMod\) is the learning rate for Kronecker factors and \(\tilde{\omega}\) is the learning rate for the diagonal re-scaling factor.

We introduce an eigenvalue corrected matrix-variate Gaussian (EMVG) posterior shown in Figure~\ref{fig:cartoon-fig1}. An EMVG is a generalization of a multivariate Gaussian distribution with the following form:
\begin{equation}
\mathcal{E}\mathcal{M}\mathcal{N}(\mathbf{W}; \mathbf{M}, \mathbf{\Sigma}_1, \mathbf{\Sigma}_2, \mathbf{R}) = \mathcal{N}(\textrm{vec}(\mathbf{W}); \textrm{vec}(\mathbf{M}), (\mathbf{Q}_{\mathbf{\Sigma}_1} \otimes \mathbf{Q}_{\mathbf{\Sigma}_2}) \mathbf{R} (\mathbf{Q}_{\mathbf{\Sigma}_1} \otimes \mathbf{Q}_{\mathbf{\Sigma}_2})^\transpose)  
\end{equation}
An EMVG posterior is potentially powerful because it not only compactly represents covariances between weights but also computes a full diagonal variance in K-FAC eigenbasis. Applying EK-FAC approximation into equation~\eqref{eq:posterior-variance} yields an EMVG posterior. Therefore, we factorize the covariance matrix in the same sense EK-FAC approximates the Fisher matrix:
\begin{equation}
\begin{aligned}
    \cov {}
    & = \frac{\lambda}{N}\left(\mathbf{Q}_{\SMat} \otimes \mathbf{Q}_{\AMat}\right) \left(\mathbf{R}^{\gamma}\right)^{-1} \left(\mathbf{Q}_{\SMat} \otimes \mathbf{Q}_{\AMat}\right)^\transpose \\
    & = \frac{\lambda}{N} (\mathbf{Q}_{\SMat} \otimes \mathbf{Q}_{\AMat}) \left(\mathbf{R} + \frac{\lambda}{N \eta}\mathbf{I}\right)^{-1} (\mathbf{Q}_{\SMat} \otimes \mathbf{Q}_{\AMat})^\transpose
\end{aligned}
\label{eq:sigma-decomp}
\end{equation}
where \(\gamma\) is an intrinsic damping term. Since the damping \(\frac{\lambda}{N \eta}\mathbf{I}\) does not affect K-FAC eigenbasis, we explicitly represent the damping term in the re-scaling matrix. In practice, it may be advantageous to add extrinsic damping to the re-scaling matrix for the stable training process.

The only difference from standard EK-FAC is that the weights are sampled from the variational posterior \(q\). We can interpret noisy EK-FAC in the sense that \(\mean\) is a point estimate of the weights and \(\cov\) is the covariance of correlated Gaussian noise for each training examples. The full algorithm is described in alg.~\ref{alg:noisy-kfac}.

The inference is efficient because the covariance matrix is factorized with three small matrices  \(\AMat, \SMat\), and \(\mathbf{R}\). We can use the following identity to compute Kronecker products efficiently: \((\mathbf{A} \otimes \mathbf{B})\textrm{vec}(\mathbf{X}) = \textrm{vec}(\mathbf{B} \mathbf{X} \mathbf{A}^\transpose)\). 

\section{Related Work}
Variational inference was first applied to neural networks by \citet{peterson1987mean} and \citet{hinton1993keeping}. Then, \citet{graves2011practical} proposed a practical method for variational inference with fully factorized Gaussian posteriors which uses a simple (but biased) gradient estimator. Improving on this work, \citet{blundell2015weight} proposed an unbiased gradient estimator using the reparameterization trick of \citet{kingma2013auto}. Several non-Gaussian variational posteriors have also been proposed such as Multiplicative Normalizing Flows \citep{louizos2017multiplicative} and implicit distributions \citep{shi2017kernel}. Neural networks with dropout were also interpreted as BNNs \citep{gal2016dropout, gal2017concrete}.


A few recent works explored structured covariance approximations by exploiting the relationship between natural gradient and variational inference. Both~\citet{zhang2017noisy} and~\citet{khan2018fast} used a diagonal Fisher approximation in natural gradient VI, obtaining a fully factorized Gaussian posterior. \citet{zhang2017noisy} also proposed an interesting extension by using K-FAC, which leads to a matrix-variate Gaussian posterior. Concurrently, \citet{mishkin2018slang} adopted a "diagonal plus low-rank" approximation. This method shares the same spirit as this work. However, their low rank approximation is computationally expensive and thus only applied to two-layer (shallow) neural networks.


\section{Experiments}
In order to empirically evaluate the proposed method, we test under two scenarios, regression and classification, to investigate the following questions. (1) Does noisy EK-FAC have improved prediction performance compared to existing methods? (2) Is it able to scale to large dataset and modern convolution neural network architecture? 
\subsection{Regression}
\begin{wrapfigure}[15]{R}{0.5\textwidth}
    \vspace{-2.0em}
    \centering
    \includegraphics[width=1.0\linewidth]{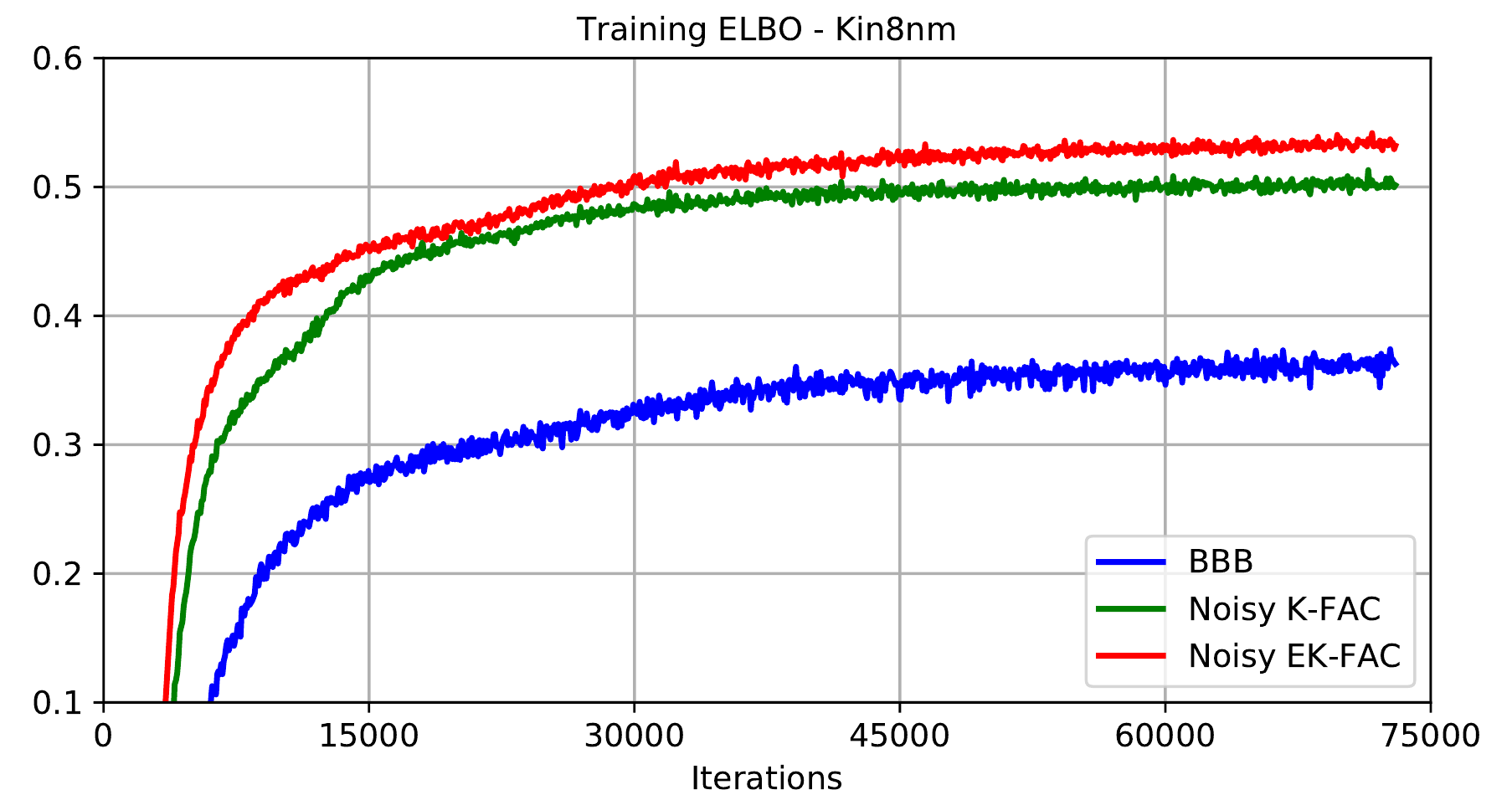}
    \small
    \vspace{-1.0em}
    \caption{Training curve for all three methods. Note that BBB, noisy K-FAC, and noisy EK-FAC use FFG, MVG, EMVG accordingly. EMVG has the most flexible variational posterior distribution.}
    \label{fig:convergence}
\end{wrapfigure}
We evaluate our method on standard BNN benchmarks from UCI collections~\citep{Dua:2017}, adopting evaluation protocols from~\citet{hernandez2015probabilistic}. In particular, we introduce a Gamma prior \(p(\gamma) = \text{Gam}(a_0 = 6, b_0 = 6)\) for the precision of Gaussian likelihood and include a Gamma posterior \(p(\gamma) = \text{Gam}(\alpha^{\gamma}, \beta^{\gamma})\) into the variational objective.

\label{experiments}
\begin{table*}[t]
\caption{Average RMSE and log-likelihood in test data for UCI regression benchmarks.}
\vspace{-0.2em}
\label{uci regression}
\begin{center}
\resizebox{\textwidth}{!}{
\begin{sc}
\begin{tabular}{lcccccccc}
\toprule
\textbf{}        & \multicolumn{4}{c}{Test RMSE} & \multicolumn{4}{c}{Test log-likelihood} \\
\textbf{Dataset} & BBB & Noisy Adam & Noisy K-FAC & Noisy EK-FAC & BBB & Noisy Adam & Noisy K-FAC & Noisy EK-FAC \\
\midrule
Boston  		 & 3.171$\pm$0.149 & 3.031$\pm$0.155 & 2.742$\pm$0.015 & \textbf{2.527$\pm$0.158} & -2.602$\pm$0.031 & -2.558$\pm$0.032 & -2.409$\pm$0.047 & \textbf{-2.378$\pm$0.044} \\
Concrete         & 5.678$\pm$0.087 & 5.613$\pm$0.113 & 5.019$\pm$0.127 & \textbf{4.880$\pm$0.120} & -3.149$\pm$0.018 & -3.145$\pm$0.023 & -3.039$\pm$0.025 & \textbf{-3.002$\pm$0.025} \\
Energy           & 0.565$\pm$0.018 & 0.839$\pm$0.046 & \textbf{0.485$\pm$0.019} & 0.497$\pm$0.023 & -1.500$\pm$0.006 & -1.629$\pm$0.020 & \textbf{-1.421$\pm$0.004} & -1.448$\pm$0.004\\
Kin8nm           & 0.080$\pm$0.001  & 0.079$\pm$0.001 & \textbf{0.076$\pm$0.001} & \textbf{0.076$\pm$0.000} & 1.111$\pm$0.007 & 1.112$\pm$0.008 & 1.148$\pm$0.007 & \textbf{1.149$\pm$0.012} \\
Naval            & \textbf{0.000$\pm$0.000} & 0.001$\pm$0.000 & \textbf{0.000$\pm$0.000} & \textbf{0.000$\pm$0.000} & 6.143$\pm$0.032 & 6.231$\pm$0.041 & 7.079$\pm$0.034 & \textbf{7.287$\pm$0.002} \\
Pow. Plant       & 4.023$\pm$0.036 & 4.002$\pm$0.039 & \textbf{3.886$\pm$0.041} & 3.895$\pm$0.053 & -2.807$\pm$0.010 & -2.803$\pm$0.010 & -2.776$\pm$0.012 & \textbf{-2.774$\pm$0.012} \\
Protein          & 4.321$\pm$0.017 & 4.380$\pm$0.016 & 4.097$\pm$0.009 & \textbf{4.042$\pm$0.027} & -2.882$\pm$0.004 & -2.896$\pm$0.004 & -2.836$\pm$0.002 & \textbf{-2.819$\pm$0.007} \\
Wine             & 0.643$\pm$0.012 & 0.644$\pm$0.011 & 0.637$\pm$0.011 & \textbf{0.635$\pm$0.013} & -0.977$\pm$0.017  & -0.976$\pm$0.016 & -0.969$\pm$0.014 & \textbf{-0.964$\pm$0.002}\\
Yacht            & 1.174$\pm$0.086 & 1.289$\pm$0.069 & 0.979$\pm$0.077 & \textbf{0.974$\pm$0.116} & -2.408$\pm$0.007 & -2.412$\pm$0.006 & -2.316$\pm$0.006 & \textbf{-2.224$\pm$0.007} \\
Year             & 9.076$\pm$NA & 9.071$\pm$NA & 8.885$\pm$NA & \textbf{8.642$\pm$NA} & -3.614$\pm$NA  & -3.620$\pm$NA & -3.595$\pm$NA & \textbf{-3.573$\pm$NA} \\ 
\bottomrule
\end{tabular}
\end{sc}
}
\end{center}
\end{table*}

We randomly split training (90\%) and test (10\%) data. To reduce the randomness, we repeat the splitting 10 times, except for two largest datasets. "Year" and "Protein" are repeated 1 and 5 times. During training, we normalize input features and training targets to zero mean and unit variance. We do not adopt this normalization at test time. All experiments train a network with a single hidden layer with 50 units except for "Protein" and "Year" datasets, which have 100 hidden units. We use batch size of 10 for smaller datasets, 100 for larger datasets, and 500 for "Year" dataset. To stabilize the training, we re-initialize the re-scaling matrix every 50 iterations with K-FAC eigenvalues. This is equivalent to executing a K-FAC iteration. We found that the second moment of the gradient vector is unstable for a small batch. We amortize the basis update to ensure re-scaling matrix matches the eigenbasis, setting \(T_{\rm eig} = 5\). For learning rates, we use  \(\alpha = 0.01\), \(\beta = 0.001\), and \(\omega = 0.01\) for all datasets. They are decayed by 0.1 for the second half of the training.

We compare the results with Bayes by Backprop~\citep{blundell2015weight}, noisy Adam, and noisy K-FAC~\citep{zhang2017noisy}. We report root mean square error (RMSE) and log-likelihood on the test dataset. The results are summarized in Table~\ref{uci regression}. The evaluation result shows that noisy EK-FAC yields a higher test log-likelihood compared to other methods. The training plot is shown in Figure~\ref{fig:convergence}. Noisy EK-FAC also achieves a higher ELBO compared to noisy K-FAC.

\subsection{Classification}
\begin{wraptable}[11]{R}{0.5\textwidth}
\vspace{-1.2em}
\caption{Classification accuracy on CIFAR10 with modified VGG16. {\bf [D]} denotes data augmentation and {\bf [B]} denotes Batch Normalization.}
\small
\vspace{-0.5em}
\label{tab:classification}
\begin{center}
\begin{tabular}{l|c|c|c|c}
\toprule
\multirow{2}{*}{Method} & \multicolumn{4}{c}{Test Accuracy}  \\ 
\cline{2-5}
&  &  D & B &  D + B \\
\midrule
SGD         & 81.79 & 88.35 & 85.75 & 91.39 \\
KFAC        & 82.39 & 88.89 & 86.86 & 92.13 \\
Noisy-KFAC  & 85.52 & 89.35 & 88.22 & 92.01 \\
Noisy-EKFAC & \textbf{87.07} & \textbf{89.86} & \textbf{88.45} & \textbf{92.22} \\ 
\bottomrule
\end{tabular}
\end{center}
\vspace{-1em}
\end{wraptable}
To evaluate the scalability of the proposed method, we train a modified VGG16~\citep{simonyan2014very} and test on CIFAR10 benchmarks~\citep{krizhevsky2009learning}. The modified VGG16 has a half reduced number of hidden units in all layers. Similar to applying K-FAC on convolutional layers with Kronecker factors~\citep{grosse2016kronecker}, EK-FAC can be extended to convolutional layers. We compare the results with SGD (with momentum), K-FAC, and noisy K-FAC. 

We use batch size of 128 for all experiments. To reduce the computational overhead, we amortize covariance, inverse, and re-scaling updates. Specifically, we set \(T_{\rm stats} = 10\), \(T_{\rm scale} = 10\), and \(T_{\rm eig} = 200\). We noticed that the amortization does not significantly impact per-iteration optimization performance. \(\beta\) and \(\omega\) are set to 0.01 for both noisy K-FAC and noisy EK-FAC. We adopt batch normalization~\citep{ioffe2015batch} and data augmentation. We tune regularization parameter \(\lambda\) and prior variance \(\eta\). With data augmentation, we use a smaller regularization parameter. \(\eta\) is set to 0.1 without batch normalization and 1.0 with batch normalization.


The results are summarized in table \ref{tab:classification}. Noisy EK-FAC achieves the highest test accuracy in all settings without introducing computational overhead. Without extra regularization tricks, noisy EK-FAC has 1.55\(\%\) improvement compared to noisy K-FAC.

\section{Conclusion}
In this paper, we introduced a modified training method for variational Bayesian neural networks. An eigenvalue corrected matrix-variate Gaussian extends on a matrix-variate Gaussian to represent the posterior distribution with more flexibility. It not only efficiently captures correlations between weights but also computes an accurate diagonal variance under K-FAC eigenbasis. For both regression and classification evaluations, noisy EK-FAC achieves higher ELBO and test accuracy, demonstrating its effectiveness.

\bibliography{nips_2018}

\appendix
\newpage

\section{Matrix-Variate Gaussian}
A matrix-variate Gaussian distribution models a multivariate Gaussian distribution for a matrix \(\mathbf{W} \in \mathbb{R}^{n \times p}\). 
\begin{equation}
    \centering
    p(\mathbf{W}| \mathbf{M}, \mathbf{U}, \mathbf{V}) = \frac{\textrm{exp}(-\frac{1}{2}\textrm{tr}[\mathbf{V}^{-1}(\mathbf{W} - \mathbf{M})^\transpose\mathbf{U}^{-1}(\mathbf{W} - \mathbf{M})])}{(2\pi)^{np/2}|\mathbf{V}|^{n/2}|\mathbf{U}|^{p/2}}
\end{equation}
where \(\mathbf{M} \in \mathbb{R}^{n \times p}\) is the mean, \(\mathbf{U} \in \mathbb{R}^{n \times n}\) is the covariance among rows and \(\mathbf{V} \in \mathbb{R}^{p \times p}\) is the covariance among columns. Since \(\mathbf{U}\) and \(\mathbf{V}\) are covariance matrices, they are positive definite. Vectorization of \(\mathbf{W}\) forms a multivariate Gaussian distribution whose covariance matrix \(\mathbb{\cov}\) is a Kronecker product of \(\mathbf{V}\) and \(\mathbf{U}\).
\begin{equation}
    \centering
    \textrm{vec}(\mathbf{W}) \sim \mathcal{N}(\textrm{vec}(\mathbf{M}), \mathbf{V} \otimes \mathbf{U}) 
\end{equation}
\section{Eigenvalue Corrected Matrix-Variate Gaussian}
An eigenvalue corrected matrix-variate Gaussian is an extension of a matrix-variate Gaussian to consider the full diagonal variance in Kronecker-factored eigenbasis. 
\begin{equation}
\begin{aligned}
    p(\mathbf{W}| \mathbf{M}, \mathbf{U}, \mathbf{V},\mathbf{R}) = \frac{\textrm{exp}(-\frac{1}{2}\mathbf{P}^\transpose \mathbf{R}^{-1}\mathbf{P})}{(2\pi)^{np/2}|\mathbf{R}|^{np/2}}~\textrm{and}~
    \mathbf{P} = \textrm{vec}( \mathbf{Q}_{\mathbf{U}}^\transpose(\mathbf{W} - \mathbf{M})\mathbf{Q}_{\mathbf{V}})
\end{aligned}
\label{eq:emvg}
\end{equation}
\(\mathbf{R} \in \mathbb{R}^{np \times np}\) is the re-scaling matrix. Because covariance matrices are positive definite, diagonal entries in \(\mathbf{R}\) are all positive. Similar to a matrix-variate Gaussian distribution, vectorization of \(\mathbf{W}\) generalizes a multivariate distribution whose covariance matrix has a Kronecker structure.
\begin{equation}
    \textrm{vec}(\mathbf{W}) \sim \mathcal{N}(\textrm{vec}(\mathbf{M}),
    \left(\mathbf{Q}_{\mathbf{V}} \otimes \mathbf{Q}_{\mathbf{U}}\right) \mathbf{R} \left(\mathbf{Q}_{\mathbf{V}} \otimes \mathbf{Q}_{\mathbf{U}}\right)^\transpose)
\end{equation}
Sampling from an eigenvalue corrected matrix-variate distribution is also a special case of sampling from a multivariate Gaussian distribution.  Let \(\mathbf{X} \in \mathbb{R}^{n \times p}\) be a matrix with independent samples for a standard multivariate Gaussian.
\begin{equation}
    \centering
    \mathbf{X} \sim \mathcal{N}(\mathbf{0}, \ident) 
\end{equation}
Then let 
\begin{equation}
    \centering
    \mathbf{Y} = \mathbf{M} + \mathbf{Q}_{\mathbf{U}} [ \mathbf{X} \odot \textrm{unvec}(\sqrt{\mathbf{r}})]\mathbf{Q}_{\mathbf{V}}^\transpose
\end{equation}
where \(\sqrt{\mathbf{r}} = \sqrt{\textrm{diag}(\mathbf{R})}\), \(\odot\) is an element-wise multiplication, and unvec\((\cdot)\) is an inverse of vec\((\cdot)\) operation.

\section{Derivation of EK-FAC Update}
\label{app:derivation}
Let \(\mathbf{V} \in \mathbb{R}^{n \times p}\) be the weight gradient, \(\mathbf{A} \in \mathbb{R}^{n \times n}\) the covariance matrix of input activations, and \(\mathbf{S} \in \mathbb{R}^{p \times p}\) the covariance matrix of output pre-activations. \(\mathbf{R} \in \mathbb{R}^{np \times np}\) is the diagonal re-scaling matrix in K-FAC eigenbasis: \(\mathbf{Q}_{\SMat} \otimes \mathbf{Q}_{\AMat}\). The following is the derivation of EK-FAC update shown in  alg.~\ref{alg:noisy-kfac}. 
\begin{equation}
\begin{aligned}
    \left(\mathbf{Q}_{\SMat} \otimes \mathbf{Q}_{\AMat}\right) \mathbf{R}^{-1} \left(\mathbf{Q}_{\SMat} \otimes \mathbf{Q}_{\AMat}\right)^\transpose \textrm{vec}(\mathbf{V}) &{} =  \left(\mathbf{Q}_{\SMat} \otimes \mathbf{Q}_{\AMat}\right) \mathbf{R}^{-1} \left(\mathbf{Q}_{\SMat}^\transpose \otimes \mathbf{Q}_{\AMat}^\transpose\right) \textrm{vec}(\mathbf{V}) \\
    & = \left(\mathbf{Q}_{\SMat} \otimes \mathbf{Q}_{\AMat}\right) \mathbf{R}^{-1} \textrm{vec}\left(\mathbf{Q}_{\AMat}^\transpose \mathbf{V}\mathbf{Q}_{\SMat}\right) \\
    & = \left(\mathbf{Q}_{\SMat} \otimes \mathbf{Q}_{\AMat}\right) \textrm{vec}((\mathbf{Q}_{\AMat_l}^\transpose\mathbf{V}_l \mathbf{Q}_{\SMat_l}) \oslash \textrm{unvec}(\mathbf{r}_l)) \\
    & = \textrm{vec}(\mathbf{Q}_{\AMat}[
        (\mathbf{Q}_{\AMat}^\transpose\mathbf{V} \mathbf{Q}_{\SMat}) \oslash \textrm{unvec}(\mathbf{r})]\mathbf{Q}_{\SMat}^\transpose)\\
\end{aligned}
\label{eq:sigma-decomp}
\end{equation}
where \(\mathbf{r} = \textrm{diag}(\mathbf{R})\), \(\oslash\) is an element-wise division, and unvec\((\cdot)\) is an inverse of vec\((\cdot)\) operation.

\section{Pseudo-Code for Noisy K-FAC}
\label{app:noisy-kfac}
\begin{algorithm}[H]
\caption{Noisy K-FAC.
A subscript \(l\) denotes the index of a layer, \(\weights_l = \kvec(\mathbf{W}_l)\), and \(\mean_l = \kvec(\mathbf{M}_l)\). We assume zero momentum for simplicity. Differences from standard K-FAC are shown in {\color{blue} blue}.}
\begin{algorithmic}
\REQUIRE \(\alpha\): stepsize
\REQUIRE \(\beta\): exponential moving average parameter for covariance factors
\REQUIRE \(\klWeight, \eta, \gamma_{\text{ex}}:\) KL weighting, prior variance, extrinsic damping term
\REQUIRE \(T_{\rm stats}, T_{\rm inv}\): stats and inverse update intervals.
\STATE \(k \leftarrow 0\) and initialize \(\{\mean_l\}_{l=1}^{L}, \{\mathbf{S}_l\}_{l=1}^{L}, \{\mathbf{A}_l\}_{l=0}^{L-1}, \{\mathbf{R}_l\}_{l=1}^{L}\)
\STATE Calculate the intrinsic damping term \({\color{blue} \gamma_{\text{in}}=\frac{\klWeight}{\ndata \eta}}\), total damping term \({\color{blue} {\gamma=\gamma_{\text{in}}+\gamma_{\text{ex}}}}\)
\WHILE{stopping criterion not met}
	\STATE \(k \leftarrow k+1\)
        \STATE {\color{blue} \(\mathbf{W}_l \sim \matrixNormal(\mathbf{M}_l,\ \frac{\klWeight}{N} [\mathbf{A}_l^{\gamma_{\text{in}}}]^{-1},\ [\mathbf{S}_l^{\gamma_{\text{in}}}]^{-1})\)}
        \IF {$k \equiv 0$ (\textrm{mod} $T_{\rm stats}$)}
	    \STATE Update the covariance factors
	    $\{\mathbf{S}_l\}_{l=1}^{L}, \{\mathbf{A}_l\}_{l=0}^{L-1}$ using eq.~\eqref{eq:moving-average}
        \ENDIF
        \IF {$k \equiv 0$ (\textrm{mod} $T_{\rm inv}$)}
	    \STATE Compute the inverses $\{[\mathbf{S}_l^\gamma]^{-1}\}_{l=1}^{L}, \{[\mathbf{A}_l^\gamma]^{-1}\}_{l=0}^{L-1}$ using eq.~\eqref{noisy-kfac}. 
        \ENDIF
        \STATE \(\mathbf{V}_l = \nabla_{\mathbf{W}_l}\log p(\target | \inputVec, \weights) {\color{blue} \,-\, \gamma_{\text{in}} \cdot \mathbf{W}_l}\)
        \STATE \(\mathbf{M}_l \gets \mathbf{M}_l + \alpha [\mathbf{A}_l^{\gamma}]^{-1} \mathbf{V}_l [\mathbf{S}_l^{\gamma}]^{-1}\)
\ENDWHILE
\end{algorithmic}
\end{algorithm}

\end{document}